\documentclass[conference]{./support/ieeeconf}
\usepackage{times}
\usepackage{amsmath}
\usepackage[pdftex]{graphicx}
\usepackage{subfigure}
\usepackage{amsmath,amssymb,amsopn,amstext,amsfonts}
\usepackage{cancel}
\usepackage[space]{cite}
\usepackage{pdfsync}
\usepackage{balance}
\usepackage{color}
\usepackage{algorithm}
\usepackage{algorithmic}
\usepackage{url}
\usepackage{booktabs}
\usepackage{threeparttable}
\usepackage[linkcolor=black,citecolor=black,urlcolor=black,colorlinks=true]{hyperref}
\usepackage{multirow}
\usepackage[outdir=./epstopdf/]{epstopdf}
\usepackage{graphicx}
\IEEEoverridecommandlockouts

\newcommand\inv[1]{#1\raisebox{1.15ex}{$\scriptscriptstyle-\!1$}}

\DeclareMathOperator*{\argmaxA}{arg\,max} % Jan Hlavacek
\DeclareMathOperator*{\argminA}{arg\,min}

\graphicspath{{./figure/}}
\DeclareGraphicsExtensions{.pdf,.png,.jpg,.eps}
\title{\LARGE \bf A General Optimization-based Framework for Local Odometry Estimation with Multiple Sensors}
\author{Tong Qin, Jie Pan, Shaozu Cao, and Shaojie Shen
\thanks{All authors are with the Department of Electronic and Computer Engineering,
        Hong Kong University of Science and Technology, Hong Kong, China.
        {\tt\small \{tong.qin, jie.pan, shaozu.cao\}@connect.ust.hk, eeshaojie@ust.hk}.}
}

\begin{document}

\maketitle
\thispagestyle{empty}
\pagestyle{empty}

\begin{abstract}
Nowadays, more and more sensors are equipped on robots to increase robustness and autonomous ability. 
We have seen various sensor suites equipped on different platforms, such as stereo cameras on ground vehicles, a monocular camera with an IMU (Inertial Measurement Unit) on mobile phones, and stereo cameras with an IMU on aerial robots. 
Although many algorithms for state estimation have been proposed in the past, they are usually applied to a single sensor or a specific sensor suite. 
Few of them can be employed with multiple sensor choices. 
In this paper, we proposed a general optimization-based framework for odometry estimation, which supports multiple sensor sets.
Every sensor is treated as a general factor in our framework. 
Factors which share common state variables are summed together to build the optimization problem.
We further demonstrate the generality with visual and inertial sensors, which form three sensor suites (stereo cameras, a monocular camera with an IMU, and stereo cameras with an IMU). 
We validate the performance of our system on public datasets and through real-world experiments with multiple sensors.
Results are compared against other state-of-the-art algorithms. 
We highlight that our system is a general framework, which can easily fuse various sensors in a pose graph optimization.
Our implementations are open source\footnote{https://github.com/HKUST-Aerial-Robotics/VINS-Fusion}.  

\end{abstract}

\section{Introduction}
Real-time 6-DoF (Degrees of Freedom) state estimation is a fundamental technology for robotics. 
Accurate state estimation plays an important role in various intelligent applications, such as robot exploration, autonomous driving, VR (Virtual Reality) and AR (Augmented Reality). 
The most common sensors we use in these applications are cameras.
A large number of impressive vision-based algorithms for pose estimation has been proposed over the last decades, such as \cite{klein2007parallel, ForPizSca1405, engel2014lsd, mur2015orb, engel2017direct}.
Besides cameras, the IMU is another popular option for state estimation. 
The IMU can measure acceleration and angular velocity at a high frequency, which is necessary for low-latency pose feedback in real-time applications. 
%Furthermore, inertial measurements are very effective for aggressive motion estimation. 
Hence, there are numerous research works fusing vision and IMU together, such as \cite{MouRou0704,LiMou1305,LeuFurRab1306,bloesch2015robust,mur2017visual,forster2017manifold, qin2018vins}. 
%Visual-inertial approaches achieve higher accuracy and robustness than pure visual approaches, thanks to the complementary property of visual features and inertial measurements. 
Another popular sensor used in state estimation is LiDAR.
LiDAR-based approaches \cite{zhang2014loam} achieve accurate pose estimation in a confined local environment.
Although a lot of algorithms have been proposed in the past, they are usually applied to a single input sensor or a specific sensor suite. 

Recently, we have seen platforms equipped with various sensor sets, such as stereo cameras on ground vehicles, a monocular camera with an IMU on mobile phones, stereo cameras with an IMU on aerial robots. 
However, as most traditional algorithms were designed for a single sensor or a specific sensor set, they cannot be ported to different platforms. 
Even for one platform, we need to choose different sensor combinations in different scenarios. 
Therefore, a general algorithm which supports different sensor suites is required. 
Another practical requirement is that in case of sensor failure, an inactive sensor should be removed and an alternative sensor should be added into the system quickly. 
Hence, a general algorithm which is compatible with multiple sensors is in need.

\begin{figure}
    \centering
    \includegraphics[width=0.45\textwidth]{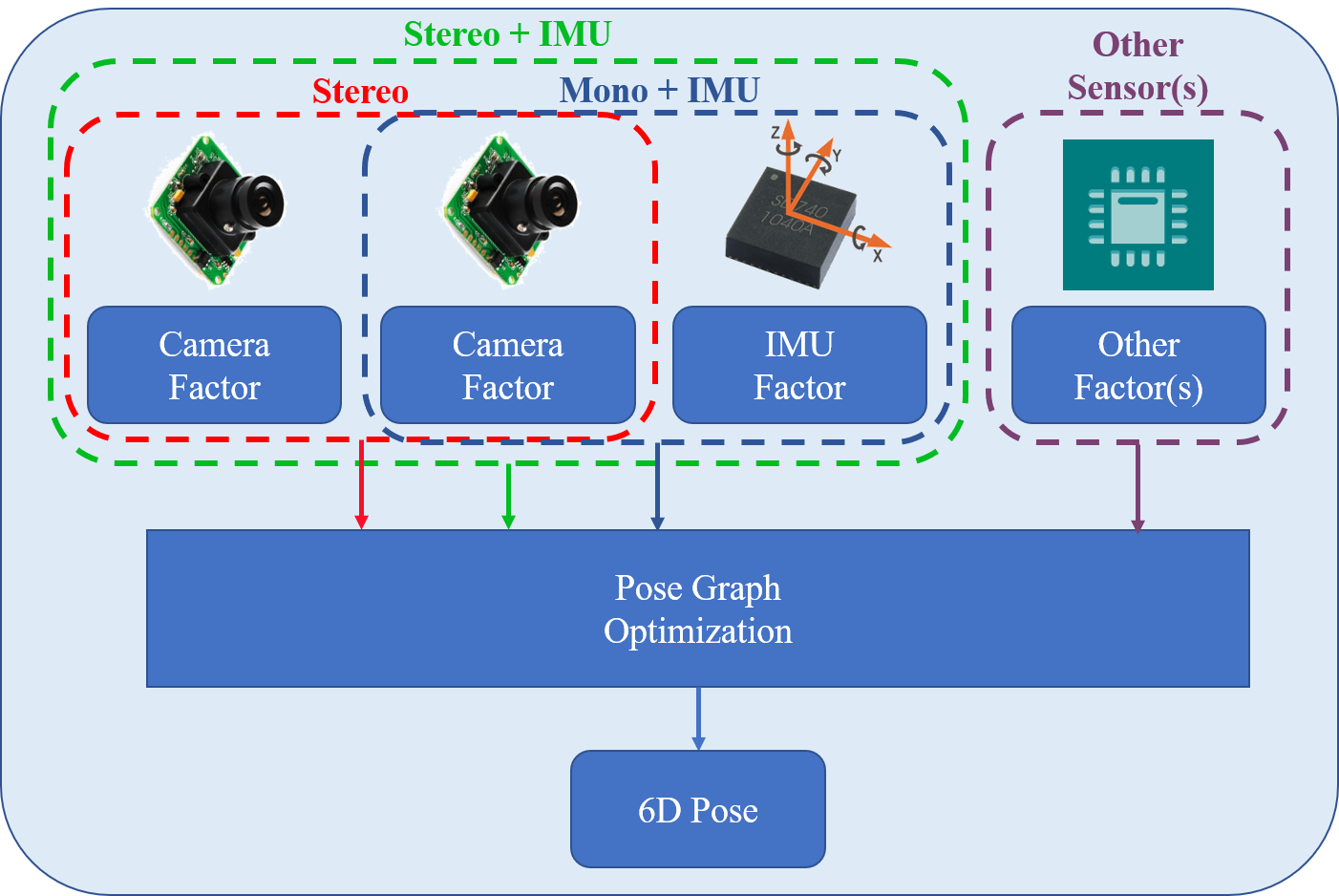}
    \caption{An illustration of the proposed framework for state estimation, which supports multiple sensor choices, such as stereo cameras, a monocular camera with an IMU, and stereo cameras with an IMU. Each sensor is treated as a general factor.
    Factors which share common state variables are summed together to build the optimization problem.
        \label{fig:framework}}
\end{figure}

In this paper, we propose a general optimization-based framework for pose estimation, which supports multiple sensor combinations.
We further demonstrate it with visual and inertial sensors, which form three sensor suites (stereo cameras, a monocular camera with an IMU, and stereo cameras with an IMU). 
We can easily switch between different sensor combinations. 
We highlight the contribution of this paper as follows:
\begin{itemize}
    \item a general optimization-based framework for state estimation, which supports multiple sensors.
    \item a detailed demonstration of state estimation with visual and inertial sensors, which form different sensor suites (stereo cameras, a monocular camera + an IMU, and stereo cameras + an IMU).
    \item an evaluation of the proposed system on both public datasets and real experiments. 
    \item open-source code for the community.
\end{itemize}

\section{Related Work}
\label{sec:literature}
State estimation has been a popular research topic over the last decades.
A large number of algorithms focus on accurate 6-DoF pose estimation.
We have seen many impressive approaches that work with one kind of sensor, such as visual-based methods \cite{klein2007parallel, ForPizSca1405, engel2014lsd, mur2015orb, engel2017direct}, LiDAR-based methods \cite{zhang2014loam}, RGB-D based methods \cite{kerl2013dense}. and event-based methods \cite{rebecq2017evo}. 
Approaches work with a monocular camera is hard to achieve 6-DoF pose estimation, since absolute scale cannot be recovered from a single camera.
To increase the observability and robustness, multiple sensors which have complementary properties are fused together.

%multiple sensor
There are two trends of approaches for multi-sensor fusion. 
One is filter-based methods, the other is optimization-based methods.
Filter-based methods are usually achieved by EKF (Extended Kalman Filter).
Visual and inertial measurements are usually filtered together for 6-DoF state estimation.
A high-rate inertial sensor is used for state propagation and visual measurements are used for the update in \cite{lynen2013robust, bloesch2015robust}.
MSCKF~\cite{MouRou0704,LiMou1305} was a popular EKF-based VIO (Visual Inertial Odometry), which maintained several camera poses and leveraged multiple camera views to form the multi-constraint update. 
Filter-based methods usually linearize states earlier and suffer from error induced by inaccurate linear points. 
To overcome the inconsistency caused by linearized error, observability constrained EKF \cite{huang2010observability} was proposed to improve accuracy and consistency. 
An UKF (Unscented Kalman Filter) algorithm was proposed in \cite{SheMulMic1405}, where visual, LiDAR and GPS measurements were fused together.
UKF is an externsion of EKF without analytic Jacobians. 
Filter-based methods are sensitive to time synchronization. 
Any late-coming measurements will cause trouble since states cannot be propagated back in filter procedure. 
Hence, special ordering mechanism is required to make sure that all measurements from multiple sensors are in order.

Optimization-based methods maintain a lot of measurements and optimize multiple variables at once, which is also known as Bundle Adjustment (BA). 
Compared with filter-based method, optimization-based method have advantage in time synchronization. 
Because the big bundle serves as a nature buffer, it can easily handle the case when measurments from multiple sensors come in disorder.
Optimization-based algorithms also outperform the filter-based algorithms in term of accuracy at the cost of computational complexity.
Early optimization solvers, such as G2O \cite{kummerle2011g}, leveraged the Gauss-Newton and Levenberg-Marquardt approaches to solve the problem.
Although the sparse structure was employed in optimization solvers, the complexity grown quadratically with the number of states and measurements.
In order to achieve real-time performance, some algorithms have explored incremental solvers, while others bounded the size of the pose graph. 
iSAM2 \cite{kaess2012isam2} was an efficient incremental solver, which reused the previous optimization result to reduce computation when new measurements came. 
The optimization iteration only updated a small part of states instead of the whole pose graph.  
Afterward, an accelerated solver was proposed in \cite{liu2018ice}, which improved efficiency by reconstructing dense structure into sparse blocks.
Methods, that keep a fixed sized of pose graph, are called sliding-window approaches.
Impressive optimization-based VIO approaches, such as~\cite{LeuFurRab1306,mur2017visual,qin2018vins}, optimized variables over a bounded-size sliding window. 
The previous states were marginalized into a prior factor without loss of information in \cite{LeuFurRab1306,qin2018vins}.
In this paper, we adopt a sliding-window optimization-based framework for state estimation.

\section{System Overview}
\label{sec:System Overview}
\begin{figure}
	\centering
	\includegraphics[width=0.45 \textwidth]{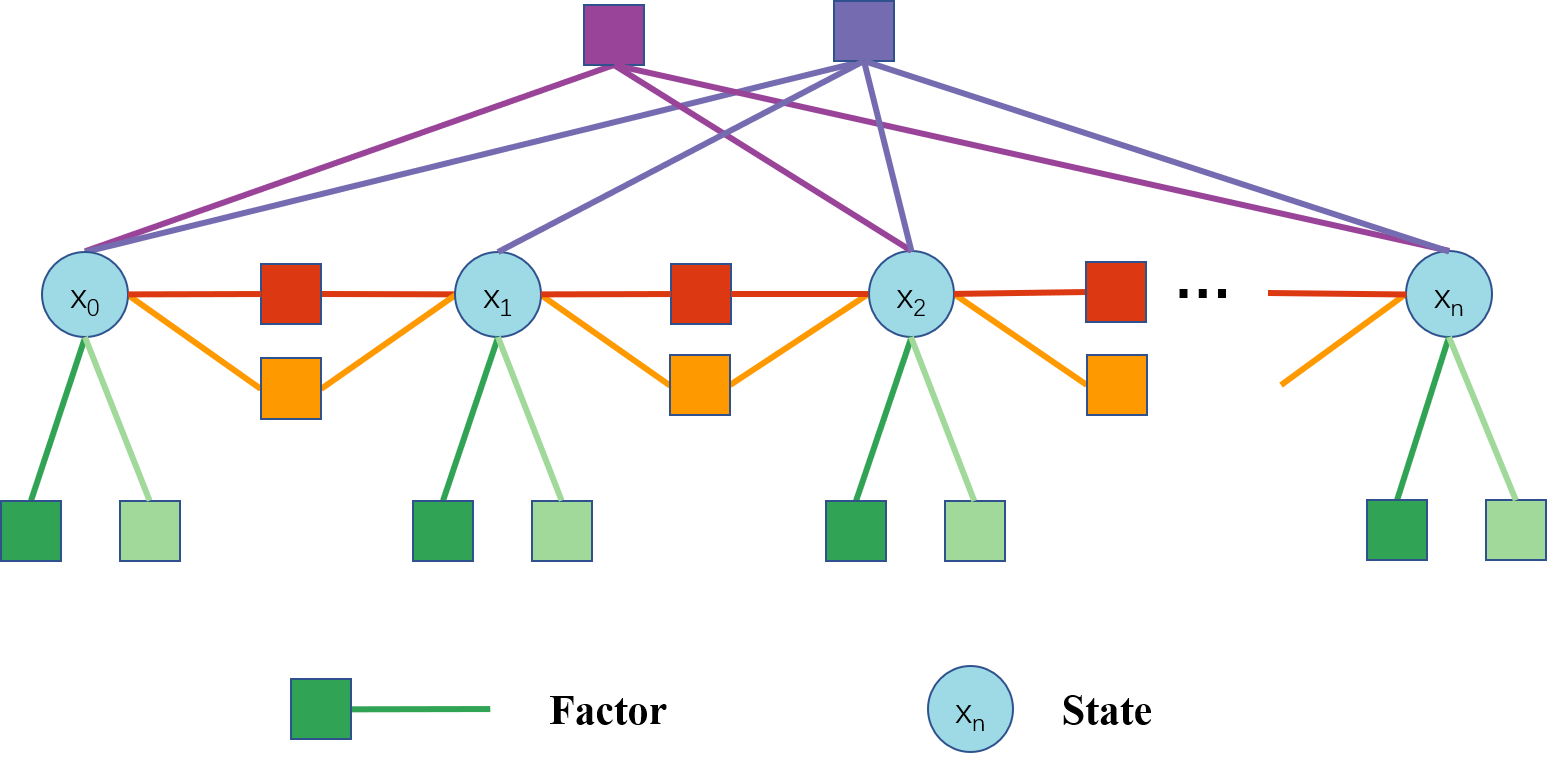}
	\caption{
		A graphic illustration of the pose graph. Each node represents states (position, orientation, velocity and so on) at one moment.
		Each edge represents a factor, which is derived by one measurement. 
		Edges constrain one state, two states or multiple states.
		\label{fig:factor_graph2}}
\end{figure}

The structure of proposed framework is shown in Fig.~\ref{fig:framework}. 
Multiple kinds of sensors can be freely combined.
The mesurement of each sensor is treated as a general factor.
Factors and their related states form the pose graph.
An illustration of pose graph is shown in Fig. \ref{fig:factor_graph2}. 
Each node represents states (position, orientation, velocity and so on) at one moment.
Each edge represents a factor, which is derived by one measurement. 
Factors constrain one state, two states or multiple states.
For IMU factor, it constrains two consecutive states by continuous motion restriction.
For a visual landmark, its factor constrains multiple states since it is observed on multiple frames. 
Once the graph is built, optimizing it equals to finding the configuration of nodes that match all edges as much as possible.

\begin{figure*}
    \centering
    \includegraphics[width=0.85 \textwidth]{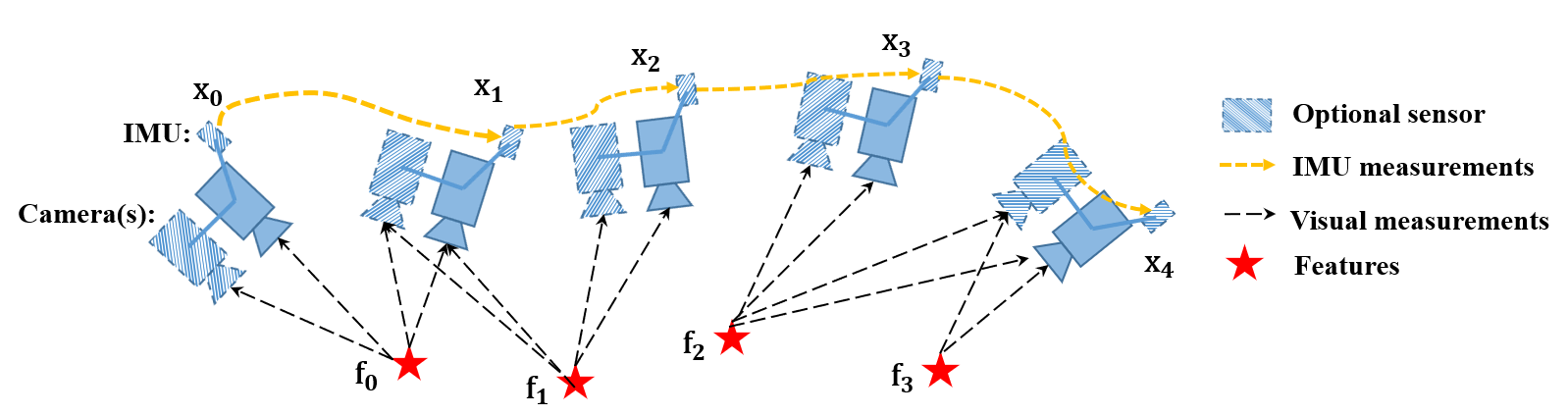}
    \caption{
        A graphic illustration of the proposed framework with visual and inertial sensors. 
        The IMU and one of cameras are optional. 
        Therefore, it forms three types (stereo cameras, a monocular camera with an IMU, and stereo cameras with an IMU).
        Several camera poses, IMU measurements and visual measurements exist in the pose graph.
        \label{fig:a_graph}}
\end{figure*}

In this paper, we specifically demonstrate the system with visual and inertial sensors.
Visual and inertial sensors can form three combinations for 6-DoF state estimation, which are stereo cameras, a monocular camera with an IMU, and stereo cameras with an IMU. 
A graphic illustration of the proposed framework with visual and inertial sensors is shown in Fig. \ref{fig:a_graph}. 
Several camera poses, IMU measurements and visual measurements exist in the pose graph.
The IMU and one of cameras are optional. 

\section{Methodology}
\label{sec:algorithm}
\subsection{Problem Definition}
\subsubsection{States}
Main states that we need to estimate includes 3D position and orientation of robot's center.
In addition, we have other optional states, which are related to sensors.
For cameras, depths or 3D locations of visual landmarks need to be estimated.
For IMU, it produces another motion variable, velocity, 
Also, time-variant acceleration bias and gyroscope bias of the IMU are needed to be estimated. 
Hence, for visual and inertial sensors, whole states we need to estimate are defined as follows:
\begin{equation}
\label{eq:variable}
\begin{split}
\mathcal{X} &=[\mathbf{p}_0, \mathbf{R}_0,\mathbf{p}_1, \mathbf{R}_1,...,\mathbf{p}_n, \mathbf{R}_n,\mathbf{x}_{cam}, \mathbf{x}_{imu}] \\
\mathbf{x}_{cam} & = [\lambda_0, \lambda_1,...,\lambda_l]\\
\mathbf{x}_{imu} & = [\mathbf{v}_0,\mathbf{b}_{a_0}, \mathbf{b}_{g_0}, \mathbf{v}_1,\mathbf{b}_{a_1}, \mathbf{b}_{g_1},... ,\mathbf{v}_n,\mathbf{b}_{a_n}, \mathbf{b}_{g_n}],
\end{split}
\end{equation}
where $\mathbf{p}$ and $\mathbf{R}$ are basic system states, which correspond to position and orientation of body expressed in world frame.
$\mathbf{x}_{cam}$ is camera-ralated state, which includes depth $\lambda$ of each feature observed in the first frame. 
$\mathbf{x}_{imu}$ is IMU-ralated variable, which is composed of  velocity $\mathbf{v}$, acceleration bias $\mathbf{b}_a$ and gyroscope bias $\mathbf{b}_g$.
$\mathbf{x}_{imu}$ can be omitted if we only use stereo camera without an IMU. 
The translation from sensors' center to body's center are assumed to be known, which are calibrated offline. 
In order to simplify the notation, we denote the IMU as body's center
(If the IMU is not used, we denote left camera as body's center).

\subsubsection{Cost Function}
The nature of state estimation is an MLE (Maximum Likelihood Estimation) problem. 
The MLE consists of the joint probability distribution of robot poses over a period of time.
Under the assumption that all measurements are independent, the problem is typically derived as,
\begin{equation}
\mathcal{X}^* = \argmaxA_{\mathcal{X}}  \prod^{n}_{t=0}\prod^{}_{k\in\mathbf{S}} p(\mathbf{z}^k_t|\mathcal{X}),
\end{equation}
where $\mathbf{S}$ is the set of measurements, which come from cameras, IMU and other sensors.
We assume the uncertainty of measurements is Gaussian distributed, $p(\mathbf{z}^k_t|\mathcal{X}) \sim \mathcal{N}(\bar{\mathbf{z}}^k_t, \Omega^k_t)$.
Therefore, the negative log-likelihood of above-mentioned equation is written as, 
\begin{equation}
\label{eq:ba}
\begin{split}
\mathcal{X}^* &= \argmaxA_{\mathcal{X}}  \prod^{n}_{t=0} \prod^{}_{k\in{\mathbf{S}}} exp(-\frac{1}{2}\left\|\mathbf{z}^k_t - h^k_t(\mathcal{X})\right\|^2_{ \mathbf{\Omega}^k_t}) \\
&= \argminA_{\mathcal{X}} \sum^n_{t=0} \sum^{}_{k\in{\mathbf{S}}}
\left\|\mathbf{z}^k_t - h^k_t(\mathcal{X})\right\|^2_{ \mathbf{\Omega}^k_t}.
\end{split}
\end{equation}
The Mahalanobis norm is defined as $\left \| \mathbf{r} \right\|_\mathbf{\Omega}^2 = \mathbf{r}^T  \mathbf{\Omega}^{-1}{\mathbf{r}}$. 
$h(\cdot)$ is the sensor model, which is detailed in the following section.
Then the state estimation is converted to a nonlinear least squares problem, which is also known as Bundle Adjustment (BA).

\subsection{Sensor Factors}
\subsubsection{Camera Factor}
The framework supports both monocular and stereo cameras. 
The intrinsic parameters of every camera and the extrinsic transformation between cameras are supposed to be known, which can be easily calibrated offline.
For each camera frame, corner features \cite{shi1994good} are detected.
These features are tracked in previous frame by KLT tracker \cite{LucKan8108}. 
For the stereo setting, the tracker also matches features between the left image and right image. 
According to the feature associations, we construct the camera factor with per feature in each frame. 
The camera factor is the reprojection process, which projects a feature from its first observation into following frames.

Considering the feature $l$ that is first observed in the image $i$, the residual for the observation in the following image $t$ is defined as:
\begin{equation}
\label{eq:vision model}
\begin{split}
\mathbf{z}^{l}_t - h^{l}_t(\mathcal{X}) &= \mathbf{z}^{l}_t - h^{l}_t (\mathbf{R}_i, \mathbf{p}_i,\mathbf{R}_t, \mathbf{p}_t, \lambda_l) \\
&=
\begin{bmatrix}
{u}^{l}_t \\
{v}^{l}_t 
\end{bmatrix} 
-\pi_c(
\inv{\mathbf{T}^b_{c}} \, \inv{ \mathbf{T}_t} \,  \mathbf{T}_i \, \mathbf{T}^b_{c} \,
\inv{\pi_c} ( \lambda_l,
\begin{bmatrix}
{u}^{l}_i \\
{v}^{l}_i 
\end{bmatrix}
)),
\end{split}
\end{equation}
where $[ {u}^{l}_i , {v}^{l}_i ] $ is the first observation of the $l$ feature that appears in the $i$ image. 
$[{u}^{l}_t ,{v}^{l}_t ] $ is the observation of the same feature in the $t$ image. 
${\pi}_c$ and ${\pi}^{-1}_c$ are the projection and back-projection functions which depend on camera model (pinhole, omnidirectional or other models). 
$\mathbf{T}$ is the 4x4 homogeneous transformation, which is $\begin{bmatrix} \mathbf{R} \,\,  \mathbf{p}\\\mathbf{0} \,\, 1 \end{bmatrix}$.
We omit some homogeneous terms for concise expression.
$\mathbf{T}^c_{b}$ is the extrinsic transformation from body center to camera center, which is calibrated offline. 
The covariance matrix $\mathbf{\Omega}^l_t$ of reprojection error is a constant value in pixel coordinate, which comes from the camera's intrinsic calibration results. 

This factor is universal for both left camera and right camera.
We can project a feature from the left image to the left image in temporal space, also we can project a feature from the left image to the right image in spatial space.
For different cameras, a different extrinsic transformation $\mathbf{T}^c_{b}$ should be used.

\subsubsection{IMU Factor}
We use the well-known IMU preintegration algorithm \cite{qin2018vins, forster2017manifold} to construct the IMU factor. 
We assume that the additive noise in acceleration and gyroscope measurements are Gaussian white noise.
The time-varying acceleration and gyroscope bias are modeled as a random walk process, whose derivative is Gaussian white noise.
Since the IMU acquires data at a higher frequency than other sensors, there are usually multiple IMU measurements existing between two frames.
Therefore, we pre-integrate IMU measurements on the manifold with covariance propagation.
The detailed preintegration can be found at \cite{qin2018vins}. 
Within two time instants, $t-1$ and $t$, the preintegration produces relative position $\boldsymbol{\alpha}^{{t-1}}_{{t}}$, velocity $\boldsymbol{\beta}^{{t-1}}_{{t}}$ and rotation $\boldsymbol{\gamma}^{{t-1}}_{{t}}$.
Also, the preintegration propagates the covariance of relative position, velocity, and rotation, as well as the covariance of bias.
The IMU residual can be defined as:
\begin{equation}
\begin{split}
&\mathbf{z}^{imu}_t - h^{imu}_t(\mathcal{X}) = 
\\
&
\begin{bmatrix}
\boldsymbol{\alpha}^{{t-1}}_{{t}}\\
\boldsymbol{\beta}^{{t-1}}_{{t}}\\
\boldsymbol{\gamma}^{{t-1}}_{{t}}\\
{\mathbf{0}}\\
{\mathbf{0}}\\
\end{bmatrix}
\ominus
\begin{bmatrix}
\inv{\mathbf{R}_{t-1}}(\mathbf{p}_{t} - \mathbf{p}_{{t-1}} + \frac{1}{2}\mathbf{g} dt^2 - \mathbf{v}_{t-1} dt) \\
\inv{\mathbf{R}_{t-1}}(\mathbf{v}_{t}  - \mathbf{v}_{{t-1}} + \mathbf{g} dt) \\
\inv{\mathbf{R}_{t-1}} \mathbf{R}_{t}\\
\mathbf{b}_{a_t} - \mathbf{b}_{a_{t-1}}\\
\mathbf{b}_{g_t} - \mathbf{b}_{g_{t-1}}
\end{bmatrix},
\end{split}
\end{equation}
where $\ominus$ is the minus operation on manifold, which is specially used for non-linear rotation. 
$dt$ is the time interval between two time instants.
$\mathbf{g}$ is the known gravity vector, whose norm is around $9.81$.
Every two adjacent frames construct one IMU factor in the cost function.

\subsubsection{Other Factors}
Though we only specify camera and IMU factors, our system is not limited to these two sensors. 
Other sensors, such as wheel speedometer, LiDAR and Radar, can be added into our system without much effort. 
The key is to model these measurements as general residual factors and add these residual factors into cost function.

\subsection{Optimization}
In traditional, the nonlinear least square problem of eq.\ref{eq:ba} is solved by Newton-Gaussian or Levenberg-Marquardt approaches. 
The cost function is linearized with respect to an initial guess of states, $\hat{\mathcal{X}}$. 
Then, the cost function is equals to:
\begin{equation}
\argminA_{\delta\mathcal{X}} \sum^n_{t=0} \sum^{}_{k\in{\mathbf{S}}}
\left\|   
\mathbf{e}^k_t + \mathbf{J}^k_t \delta\mathcal{X}
\right\|^2_{ \mathbf{\Omega}^k_t} ,
\end{equation}
where $\mathbf{J}$ is the Jacobian matrix of each factor with respect to current states $\hat{\mathcal{X}}$.
After linearization approximation, this cost function has closed-form solution of $\delta{\mathcal{X}}$.
We take Newton-Gaussian as example, the solution is derived as follows,
\begin{equation}
\label{eq:hb}
\underbrace{\sum\sum{\mathbf{J}^k_t}^T{\mathbf{\Omega}^k_t}^{-1}\mathbf{J}^k_t}_{\mathbf{H}}
\delta\mathcal{X} = 
\underbrace{-\sum\sum{\mathbf{J}^k_t}^T{\mathbf{\Omega}^k_t}^{-1}\mathbf{e}^k_t}_{\mathbf{b}}.
\end{equation}
Finally, current state $\hat{\mathcal{X}}$ is updated with $\hat{\mathcal{X}} \oplus \delta \mathcal{X}$, where $\oplus$ is the plus operation on manifold for rotation.
This procedure iterates several times until convergence. 
We adopt Ceres solver\cite{ceres-solver} to solve this problem, which utilizes advanced mathematical tools to get stable and optimal results efficiently.

\subsection{Marginalization}

Since the number of states increases along with time, the computational complexity will increase quadratically accordingly.
In order to bound the computational complexity, marginalization is incorporated without loss of useful information. 
Marginalization procedure converts previous measurements into a prior term, which reserves past information. 
The set of states to be marginalized out is denoted as $\mathcal{X}_m$, and the set of remaining states is denoted as $\mathcal{X}_r$.
By summing all marginalized factors (eq.\ref{eq:hb}), we get a new $\mathbf{H}$ and $\mathbf{b}$.
After rearrange states' order, we get the following relationship:

\begin{equation}
\begin{bmatrix}
\mathbf{H}_{mm} \, \, \mathbf{H}_{mr}\\
\mathbf{H}_{rm} \, \, \mathbf{H}_{rr}
\end{bmatrix}
\begin{bmatrix}
\delta\mathcal{X}_m\\
\delta\mathcal{X}_r
\end{bmatrix}
=
\begin{bmatrix}
\mathbf{b}_m\\
\mathbf{b}_r
\end{bmatrix}.
\end{equation}
The marginalization is carried out using the Schur complement \cite{SibMatSuk1009} as follows:

\begin{equation}
\underbrace{(\mathbf{H}_{rr}-\mathbf{H}_{rm}\inv{\mathbf{H}_{mm}}\mathbf{H}_{mr})}_{\mathbf{H}_p}
\delta\mathcal{X}_r
=
\underbrace{\mathbf{b}_r - \mathbf{H}_{rm}\inv{\mathbf{H}_{mm}}\mathbf{b}_m}_{\mathbf{b}_p}.
\end{equation}
We get a new prior $\mathbf{H}_p, \mathbf{b}_p$ for the remaining states. 
The information about marginalized states is converted into prior term without any loss.
To be specific, we keep ten spacial camera frames in our system.
When a new keyframe comes, we marginalize out the visual and inertial factors, which are related with states of the first frame.

After we get the prior information about current states,
with Bayes' rule, we could calculate the posterior as a product of likelihood and prior: $p(\mathcal{X}|\mathbf{z}) \propto p(\mathbf{z}|\mathcal{X})p(\mathcal{X})$.
The state estimation then becomes a MAP (Maximum A Posteriori) problem.
Denote that we keep states from instant m to instant n in the sliding window.
The states before m are marginalized out and converted to a prior term.
Therefore, the MAP problem is written as:
\begin{equation}
\begin{split}
\mathcal{X}_{m:n}^* &= \argmaxA_{\mathcal{X}_{m:n}}  \prod^{n}_{t=m}\prod^{}_{k\in{\mathbf{S}}} p(\mathbf{z}^k_t|\mathcal{X}_{m:n}) \, \, \,p(\mathcal{X}_{m:n}) \\
&= \argminA_{\mathcal{X}_{m:n}} \sum^n_{t=m} \sum^{}_{k\in{\mathbf{S}}}
\left\|\mathbf{z}^k_t - h^k_t(\mathcal{X}_{m:n})\right\|^2_{ \mathbf{\Omega}^k_t}  \\
& \qquad \qquad \,\,+(\mathbf{H}_p \delta \mathcal{X}_{m:n} - \mathbf{b}_p).
\end{split}
\end{equation}
Compared with eq.\ref{eq:ba}, the above-mentioned equation only adds a prior term. 
It is solved as same as eq.\ref{eq:ba} by Ceres solver\cite{ceres-solver}.

\subsection{Discussion}
The proposed system is a general framework.
Various sensors can be easily added into our system, as long as it can be derived as a general residual factor. 
Since our system is not specially designed for a certain sensor, it is capable to handle sensor failure case.
When sensor failure occurs, we just remove factors of the inactive sensor and add new factors from other alternative sensors.

%Our system failed to deal with the monocular-only case. 
%Because the absolute scale is unobservable, the 6-DoF pose cannot be recovered from a single camera. 
%A special procedure to lock the scale is required, which is not included in our framework.

%real time

\section{Experimental Results}
\label{sec:experiments}
We evaluate the proposed system with visual and inertial sensors both on datasets and with real-world experiments. 
In the first experiment, we compare the proposed algorithm with another state-of-the-art algorithm on public datasets. 
We then test our system in the large-scale outdoor environment.
The numerical analysis is generated to show the accuracy of our system in detail.

\subsection{Datasets}

\begin{figure}
	\centering
	\includegraphics[width=0.5\textwidth]{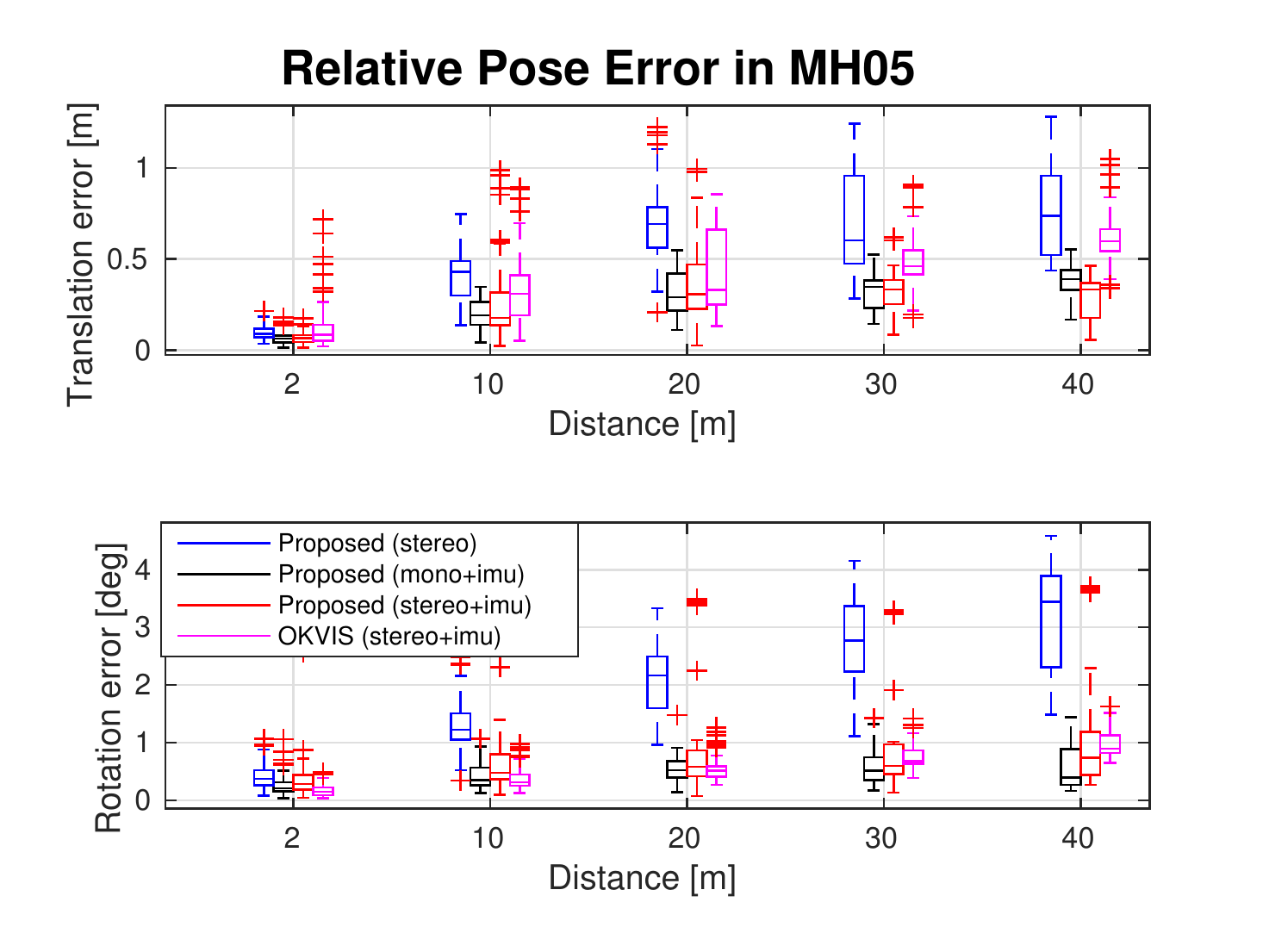}
	\caption{
		Relative pose error \cite{geiger2012we} in MH\_05\_difficult.
		Two plots are relative errors in translation and rotation respectively.
		\label{fig:mh05}}
\end{figure}

\begin{figure}
	\centering
	\includegraphics[width=0.5\textwidth]{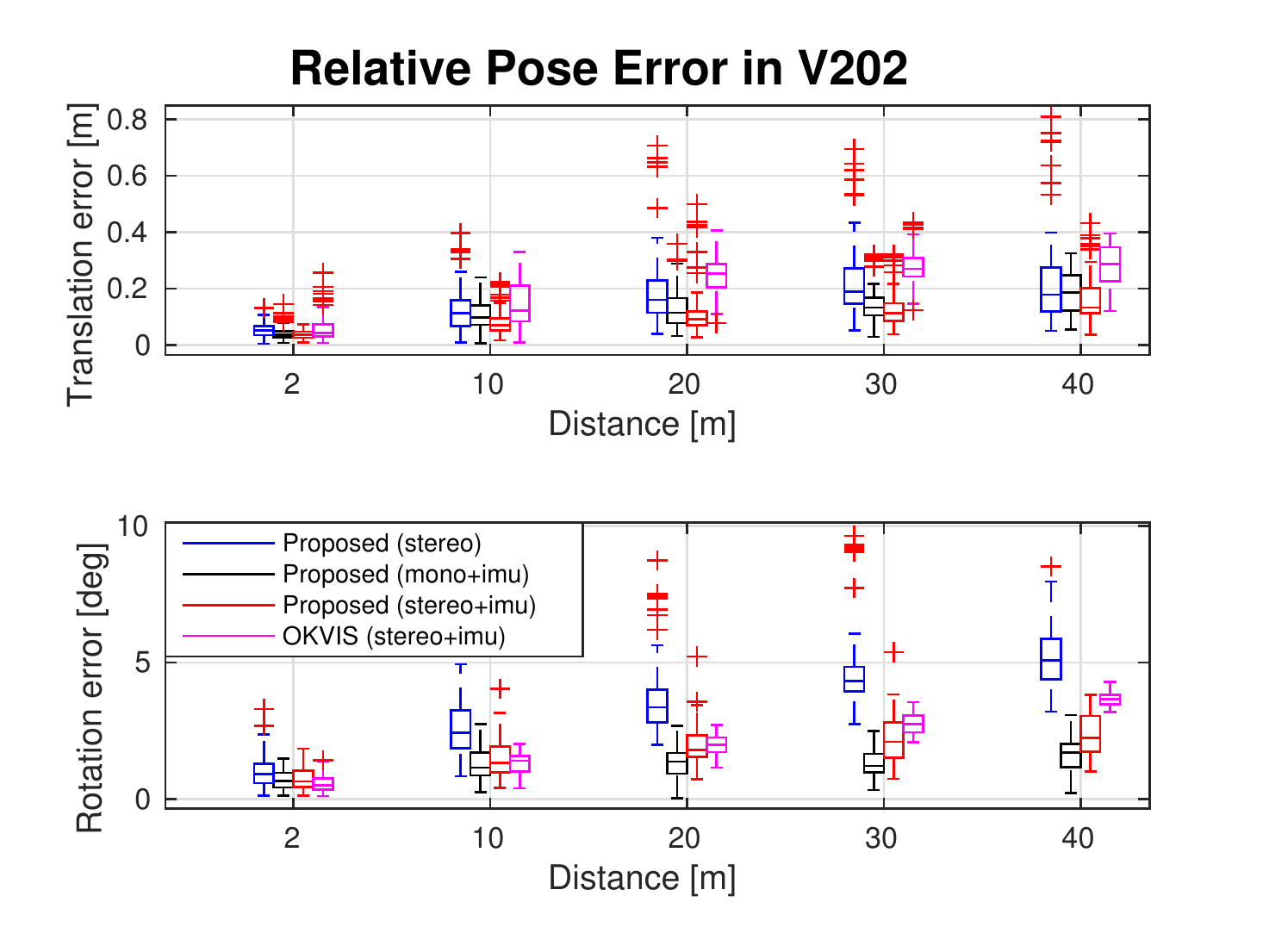}
	\caption{
		Relative pose error \cite{geiger2012we} in V2\_02\_medium.
		Two plots are relative errors in translation and rotation respectively.
		\label{fig:V202}}
\end{figure}

\begin{table}
	\centering
	\caption{{RMSE[m] in EuRoC dataset.} \label{tab:euroc}}
	\begin{tabular}{c c c c c c}
		\toprule
		\multirow{2}{*}{Sequence} & \multirow{2}{*}{Length} &
		\multicolumn{3}{c}{Proposed RMSE} &
		{OKVIS  } \\
		\cline{3-5}
		&     & stereo & mono+imu &stereo+imu & RMSE \\
		\hline
		MH\_01 & 79.84 & 0.54 & 0.18 & 0.24 & \textbf{0.16} \\
		MH\_02 & 72.75 & 0.46 & \textbf{0.09} & 0.18 & 0.22\\
		MH\_03 & 130.58 & 0.33 & \textbf{0.17} & 0.23 & 0.24 \\
		MH\_04 & 91.55 & 0.78 & \textbf{0.21} & 0.39 & 0.34\\
		MH\_05 & 97.32 & 0.50 & 0.25 & \textbf{0.19} & 0.47\\
		V1\_01 & 58.51 & 0.55 & \textbf{0.06} & 0.10 & 0.09\\
		V1\_02 & 75.72 & 0.23 & \textbf{0.09} & 0.10 & 0.20\\
		V1\_03 & 78.77 & x & 0.18 &  \textbf{0.11} & 0.24\\
		V2\_01 & 36.34 & 0.23 & \textbf{0.06} & 0.12 & 0.13\\
		V2\_02 & 83.01 &0.20 & 0.11 & \textbf{0.10} & 0.16 \\
		V2\_03 & 85.23 &x & \textbf{0.26} & 0.27 & 0.29\\
		\bottomrule
	\end{tabular}
	%\begin{tablenotes}
	\footnotesize
	%\item[1]   RMSE is root mean square error, as proposed in \cite{sturm2012benchmark}.
	%\end{tablenotes}
\end{table}

We evaluate our proposed system using the EuRoC MAV Visual-Inertial Datasets~\cite{Burri25012016}. 
This datasets are collected onboard a micro aerial vehicle, 
which contain stereo images (Aptina MT9V034 global shutter, 752x480 monochrome, 20 FPS), 
synchronized IMU measurements (ADIS16448, 200 Hz), 
Also, the ground truth states are provided by VICON and Leica MS50.
We run datasets with three different combinations of sensors, which are stereo cameras, a monocular camera with an IMU, stereo cameras with an IMU separately.

In this experiment, we compare our results with OKVIS~\cite{LeuFurRab1306}, a state-of-the-art VIO that works with stereo cameras and an IMU. 
OKVIS is another optimization-based sliding-window algorithm. 
OKVIS is specially designed for visual-inertial sensors, while our system is a more general framework, which supports multiple sensors combinations. 
We tested the proposed framework and OKVIS with all sequences in EuRoC datasets.
We evaluated accuracy by RPE (Relative Pose Errors) and ATE (Absolute Trajectory Errors). 
The RPE is calculated by tools proposed in \cite{geiger2012we}.
The RPE (Relative Pose Errors) plot of two sequences, MH\_05\_difficult and V2\_02\_medium, are shown in Fig.~\ref{fig:mh05} and Fig. \ref{fig:V202} respectively.

The RMSE (Root Mean Square Errors) of ATE for all sequences in EuRoC datasets is shown in Table.~\ref{tab:euroc}. 
Estimated trajectories are aligned with the ground truth by Horn's method \cite{horn1987closed}.
The stereo-only case fails in V1\_03\_difficult and V2\_03\_difficult sequences, where the movement is too aggressive for visual tracking to survive. 
Methods which involves the IMU work successfully in all sequences.
It is a good case to show that the IMU can dramatically improve motion tracking performance by bridging the gap when visual tracks fail due to illumination change, texture-less area, or motion blur.

From the relative pose error and absolute trajectory error, we can see that the stereo-only method performed worst in most sequences. 
Position and rotation drift obviously grown along with distance in stereo-only case.
In other words, the IMU significantly benefited vision in states estimation.
Since the IMU measures gravity vector, it can effectively suppress drifts in roll and pitch angles. 
Stereo cameras with an IMU didn't always perform best, because it requires more accurate calibration than the case of a monocular camera with an IMU.
Inaccurate intrinsic and extrinsic calibration will introduce more noise into the system. 
In general, multiple sensor fusion increase the robustness of the system.
Our results outperforms OKVIS in most sequences.

\subsection{Real-world experiment}

\begin{figure}
	\centering
	\includegraphics[width=0.4\textwidth]{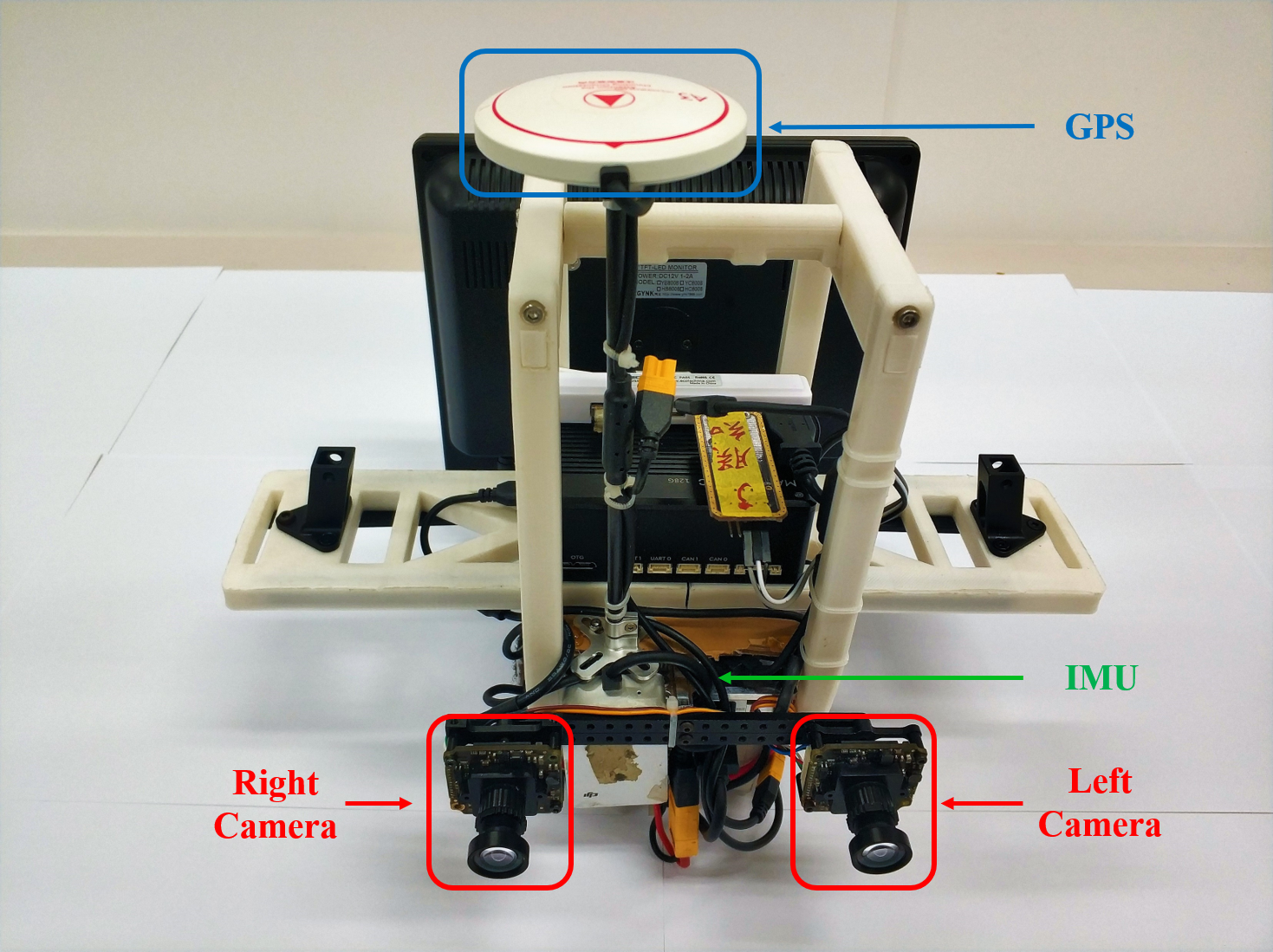}
	\caption{
		The self-developed sensor suite used in the outdoor environment. 
		It contains stereo cameras (mvBlueFOX-MLC200w, 20Hz) and DJI A3 controller, which include inbuilt IMU (200Hz) and GPS receiver. 
		\label{fig:device}}
\end{figure}

\begin{figure}
	\centering
	\includegraphics[width=0.45\textwidth]{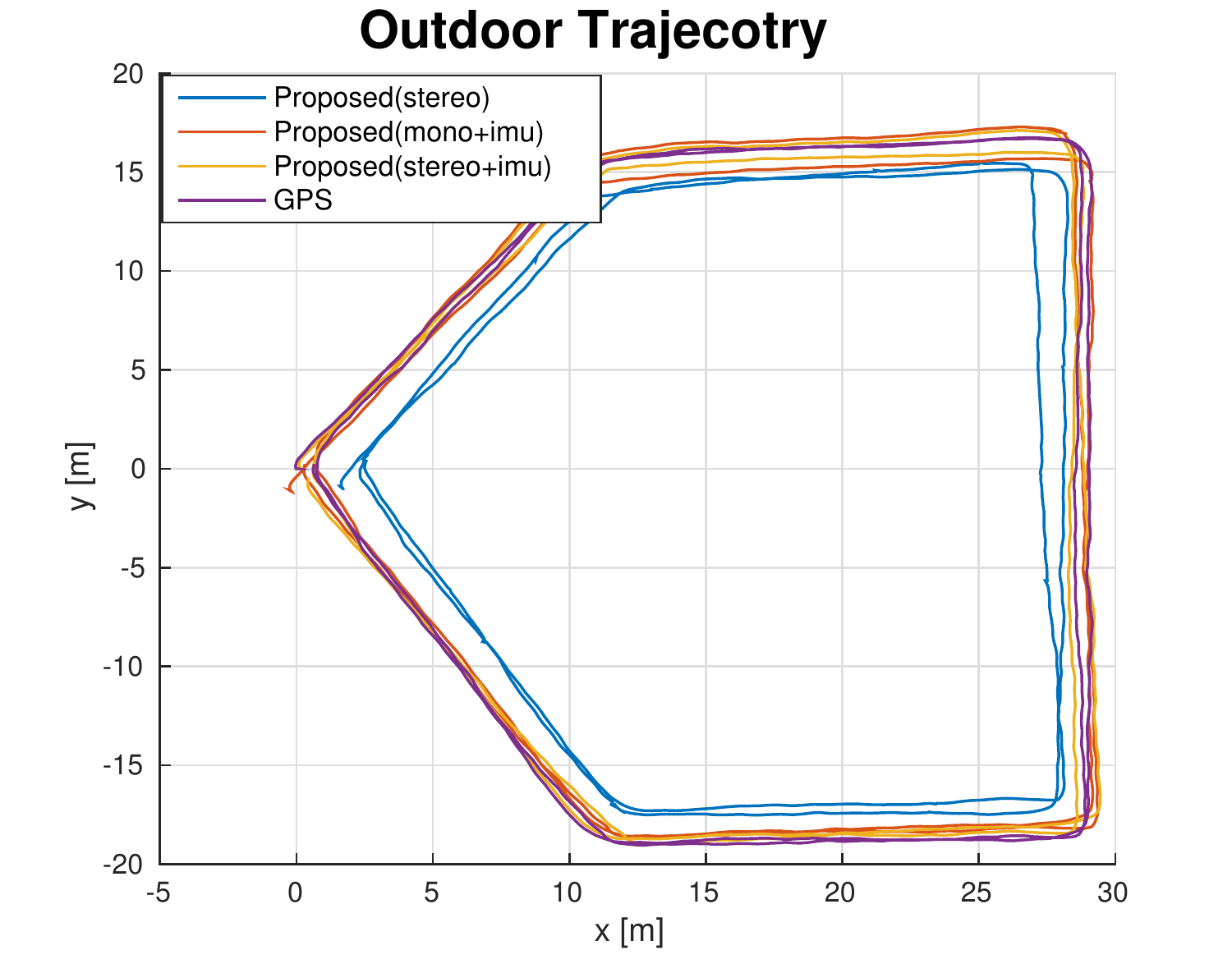}
	\caption{
		Estimated trajectories in outdoor experiment.
		\label{fig:trajectory_a}}
\end{figure}

\begin{figure}
	\centering
	\includegraphics[width=0.5\textwidth]{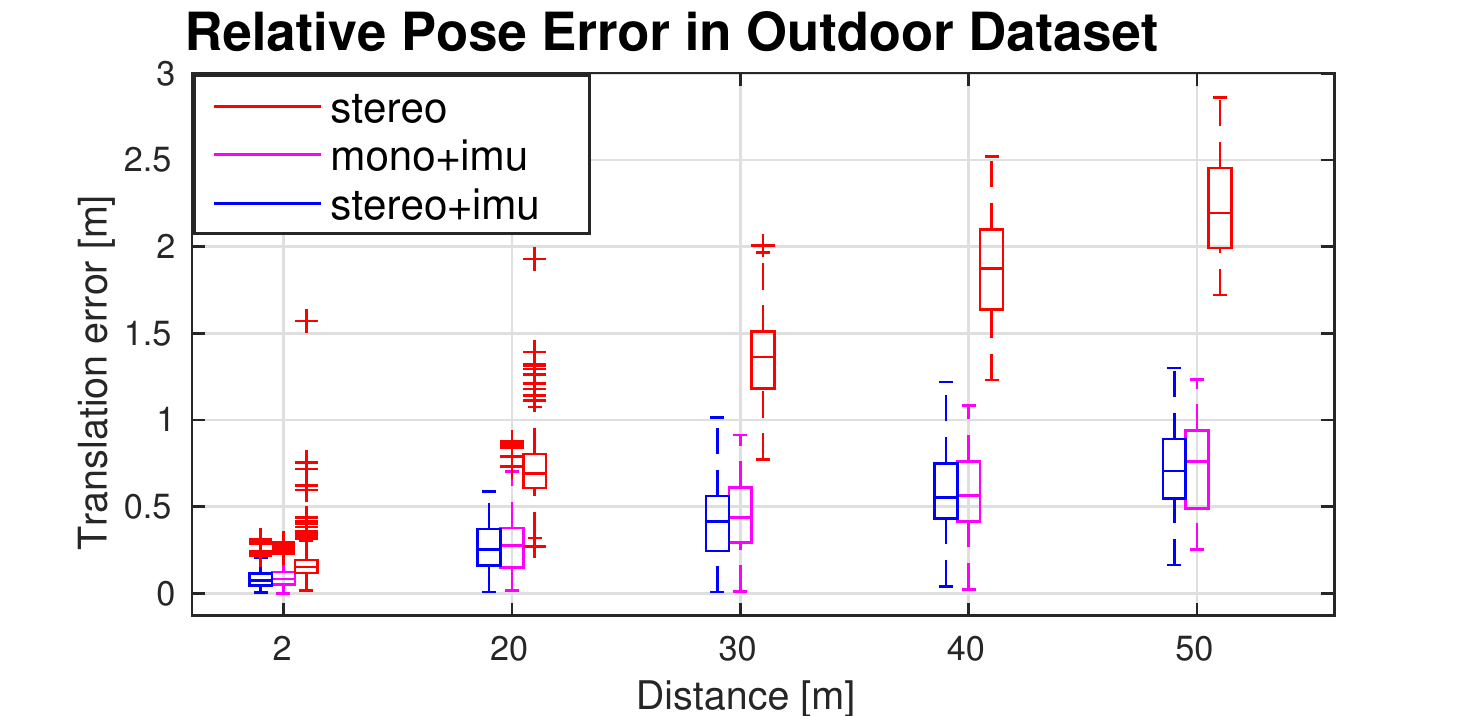}
	\caption{
		Relative pose error \cite{geiger2012we} in outdoor experiment.
		\label{fig:rpe_a}}
	\vspace{-0.3cm}
\end{figure}

\begin{table}
	\centering
	\caption{{RMSE[m] in outdoor experiment.} \label{tab:tab_outdoor}}
	\begin{tabular}{c c c c c}
		\toprule
		\multirow{2}{*}{Sequence} & \multirow{2}{*}{Length} &
		\multicolumn{3}{c}{Proposed RMSE} \\
		\cline{3-5}
		&     & stereo & mono+imu &stereo+imu\\
		\hline
		outdoor1 & 223.70 & 1.85 & 0.71 & \textbf{0.52}  \\
		outdoor2 & 229.91 & 2.35 & 0.56 & \textbf{0.43} \\
		outdoor3 & 232.13 & 2.59 & \textbf{0.65} & 0.75 \\
		\bottomrule
	\end{tabular}
	%\begin{tablenotes}
	%    \footnotesize
	%    \item[1]   RMSE is root mean square error, as proposed in \cite{sturm2012benchmark}.
	%\end{tablenotes}
\end{table}

In this experiment, we used a self-developed sensor suite to demonstrate our framework. 
The sensor suite is shown in Fig.~\ref{fig:device}.
It contains stereo cameras (mvBlueFOX-MLC200w, 20Hz) and DJI A3 controller\footnote{\url{http://www.dji.com/a3}}, which inculdes inbuilt IMU (200Hz) and GPS receiver. 
The GPS position is treated as ground truth.
We hold the sensor suite by hand and walk around on the outdoor ground. 
We run states estimation with three different combinations, which are stereo cameras, a monocular camera with an IMU, and stereo cameras with an IMU.

For accuracy comparison, we walked two circles on the ground and compared our estimation with GPS. 
The trajectory is shown in Fig.~\ref{fig:trajectory_a}, and the RPE (Relative Pose Error) is shown in Fig.~\ref{fig:rpe_a}.
As same as dataset experiment, noticeable position drifts occurred in the stereo-only scenario. 
With the assistance of the IMU, the accuracy improves a lot.
The RMSE of more outdoor experiments is shown in Table.~\ref{tab:tab_outdoor}.
The method which involves the IMU always performs better than the stereo-only case.

\section{Conclusion}
\label{sec:conclusion}

In this paper we have presented a general optimization-based framework for local pose estimation.
The proposed framework can support multiple sensor combinations, which is desirable in aspect of robustness and practicability.
We further demonstrate it with visual and inertial sensors, which form three sensor suites (stereo cameras, a monocular camera with an IMU, and stereo cameras with an IMU). 
Note that although we only show the factor formulations for the camera and IMU, our framework can be generalized to other sensors as well. 
We validate the performance of our system with multiple sensors on both public datasets and real-world experiments. 
The numerical result indicates that our framework is able to fuse sensor data with different settings.

In future work, we will extend our framework with global sensors (e.g. GPS) to achieve locally accurate and globally aware pose estimation.

\clearpage
\bibliography{paper.bib}

\end{document}